\documentclass[11pt,a4paper]{article}

\usepackage[T1]{fontenc}
\usepackage[utf8]{inputenc}
\usepackage{lmodern}
\usepackage[a4paper,margin=27mm]{geometry}
\usepackage{microtype}
\usepackage{amsmath,amssymb}
\usepackage{booktabs,tabularx,array}
\usepackage{enumitem}
\usepackage[round,authoryear]{natbib}
\usepackage{xurl,xcolor}
\usepackage{needspace}

\definecolor{linkblue}{RGB}{0,70,140}
\definecolor{citegreen}{RGB}{0,100,80}
\definecolor{urlblue}{RGB}{0,90,160}

\usepackage[
  colorlinks=true,
  linkcolor=linkblue,
  citecolor=citegreen,
  urlcolor=urlblue
]{hyperref}

\usepackage[nameinlink,capitalise]{cleveref}

\hypersetup{
  pdftitle={A Human Audit of OpenAI's AI-Generated Mathematical Proofs},
  pdfauthor={Mikolaj Sienicki and Krzysztof Sienicki}
}

\newcommand{\doi}[1]{%
  \href{https://doi.org/#1}{\nolinkurl{#1}}%
}

\newcommand{\AIGSCommunityURL}{%
  \url{https://zenodo.org/communities/ai-generated-science/records?q=&l=list&p=1&s=10&sort=newest}%
}

\newcolumntype{Y}{>{\raggedright\arraybackslash}X}

\setlist[itemize]{
  leftmargin=1.5em,
  topsep=4pt,
  itemsep=3pt,
  parsep=0pt
}

\begin{document}

\begin{center}

{\LARGE\bfseries
A Human Audit of OpenAI's\\[0.2em]
AI-Generated Mathematical Proofs
\par}

\vspace{0.8em}

{\large
Miko{\l}aj Sienicki$^{1}$ and Krzysztof Sienicki$^{2}$
\par}

\vspace{0.7em}

{\small
$^{1}$Polish-Japanese Academy of Information Technology\\
ul. Koszykowa 86, 02-008 Warsaw, Poland, European Union
\par

\vspace{0.45em}

$^{2}$Chair of Theoretical Physics of Naturally Intelligent Systems
(NIS)\footnotemark[1]\\
ul. Lipowa 2 / Topolowa 19, 05-807 Podkowa Le\'sna,
Poland, European Union
\par}

\vspace{0.65em}

\today

\end{center}

\begin{abstract}

What has human scrutiny established about the ten mathematical results
announced by OpenAI on 1 August 2026? We examine the evidence in 18
chapter-specific assessments, read alongside the project's review standards,
public Lean formalizations, subsequent research, and mathematical reference
entries. The subject is the reliability and reach of this review record;
the article does not claim a new, complete reconstruction of all ten proofs.

No surviving confirmed substantive mathematical error in a principal result
appears in the assessments examined. The qualification matters: the reports
differ in depth, and several dependencies have only been partly checked.
Chapter~8 remains the most substantial reservation, with one specialist
review calling for major revision of compressed analytic passages. In
Chapter~6, by contrast, an apparent polarity error was withdrawn after the
typeset source revealed an overbar lost during PDF extraction.

Subsequent work adds evidence of different kinds. Chapter~3 has independent
reuse of its proof mechanism; Chapter~4 has confirmation that Connes's
rigidity conjecture is false, without independent reproduction of OpenAI's
stronger infinite-family result. Chapter~7 has the strongest direct
theorem-level corroboration among the cited follow-ups through a stronger
hardness theorem. Chapter~8 has complementary work on equality, which does
not verify the analytic inequality proof. Some follow-ups disclose material
AI assistance. We argue for confidence built through formal checking,
human reconstruction, independent mathematical use, and a public record
that allows errors in both proofs and reviews to be corrected.

\end{abstract}

\section{The question behind the audit}

A collection of favorable reports can give a misleading impression of
closure. Readers need to know which arguments were reconstructed, which
dependencies were accepted from the literature, and which questions remain
open. These distinctions become especially important when a research
announcement brings together AI-generated arguments, human-readable
manuscripts, and formal proof certificates. Each may support confidence,
but each answers a different question.

OpenAI's \emph{Ten Advances in Mathematics and Theoretical Computer Science}
covers geometry, coding theory, group theory, operator algebras, arithmetic
complexity, quantum information, lattice complexity, and extremal
combinatorics. According to OpenAI's account, an internal model produced the
arguments, humans prepared the manuscripts with model assistance, and the
model subsequently formalized the results in Lean
\citep{OpenAI2026TenAdvances}. This sequence makes the connection between
the published mathematics and its formal counterpart an essential part of
the assessment.

The source record used here is fixed to the manuscript dated 1 August 2026
\citep{OpenAI2026Manuscript} and the public \texttt{ten-proofs}
repository at commit \texttt{94bc0fe}, dated 2 August 2026
\citep{OpenAI2026Lean,OpenAI2026LeanSnapshot}. That snapshot specifies
Lean~4.32.0. Its metadata list a \texttt{sorry\_count} of zero for the
principal formalized results corresponding to all ten advertised results.
Comparator or challenge files can contain deliberate \texttt{sorry}
placeholders; their presence should not be attributed to the principal
certificates.

Lean's logical kernel provides a strong check on formal derivations
\citep{deMouraUllrich2021}. What a successful check establishes is the
encoded proposition relative to the kernel, definitions, imported results,
and axioms. Human scrutiny still has to establish the intended meaning of
that proposition. It must also examine the informal account, the hypotheses
of imported theorems, and the historical claims made for the contribution.
OpenAI's earlier \emph{First Proof} exercise offers a relevant precedent:
one attempt initially considered likely correct was reassessed after expert
and community criticism \citep{OpenAI2026FirstProof}. Specialist judgment
remains necessary even when a proof first looks convincing.

Our question concerns the support supplied by the available assessments:
how much was checked, how reliable were the objections, and what has later
mathematical work added? This defines the scope of the present audit.
A human audit may examine the evidentiary basis for a proof without
claiming to have reconstructed every line of it. Equally, a criticism must
be tested against the actual manuscript before it becomes evidence of a
mathematical defect.

\section{Reading the review record}

\subsection{Coverage, responsibility, and declared standards}

The corpus examined on 27 August 2026 contains 23 \LaTeX{} documents.
Eighteen are chapter-specific referee reports, technical assessments, or
critical notes. The other five contain the audit protocol, ethical guidance,
and broader methodological material \citep{PeerReviewArchive2026}. Every
chapter is represented, although the amount and kind of attention vary.
Some assessments work through constants or lemmas; others concentrate on
proof architecture, exposition, literature dependencies, or the match to
formalized statements. Document counts therefore measure coverage, not the
number of independent reviewers or complete independent verifications.

\begin{table}[htbp]
\centering
\small

\caption{Coverage in the 27 August 2026 snapshot. Counts refer to
chapter-specific documents. The qualifications are retained separately from
the favorable overall findings.}

\label{tab:coverage}

\begin{tabularx}{\textwidth}{@{}c c Y@{}}
\toprule
Chapter & Reports & Focus and limit of the assessment \\
\midrule

1 & 2 &
Asymptotic constant, normalization, and Mellin/Fourier structure;
the analytic estimates were not all freshly derived.
\\

2 & 2 &
Binary, spherical, and packing bounds; representation-theoretic
and compactness interfaces warrant specialist attention.
\\

3 & 2 &
Non-sofic construction; conventions and nested auxiliary
parameters require careful matching.
\\

4 & 1 &
Favorable assessment of the principal construction;
comparatively little review redundancy.
\\

5 & 3 &
Favorable proof and parameter checks; novelty and attribution
require a separate judgment.
\\

6 & 1 &
The polarity objection was withdrawn after the typeset
notation was inspected.
\\

7 & 2 &
Reconstruction and exponent bookkeeping are favorable;
some threshold and degree calculations remain compressed.
\\

8 & 2 &
No confirmed error; unresolved or incompletely developed
analytic transitions prevent a claim of full verification.
\\

9 & 1 &
Recursive construction and asymptotics are judged consistent;
the corpus contains only one chapter-specific assessment.
\\

10 & 2 &
Both main results are assessed favorably;
one optimization needs an explicit derivation.
\\

\bottomrule
\end{tabularx}
\end{table}

The project asked reviewers to use the \emph{AI-Generated Science (AIGS)
Scope, Policies and Classification Standard} \citep{AIGSScope2026} and,
where applicable, the dedicated \emph{Ethical Principles and Reviewer
Assessment Form} \citep{AIGSEthics2026}. The latter was published on
15 August 2026. Earlier reports can be said to have followed its requirements
only where those requirements, or substantially equivalent principles, had
already been communicated. The project used or progressively adopted this
framework; publication of the form does not retroactively establish
compliance by every earlier report.

The framework asks reviewers to state the scope of their work, relevant
conflicts of interest, material AI assistance, and uncertainty. It separates
correctness from originality and significance, and asks for attention to
the relation between a Lean proposition and the theorem in the manuscript.
These requirements make reports interpretable across specialties while
allowing their mathematical methods to differ.

Responsibility also needs a precise meaning. We use
\emph{human-accountable assessment} for a report whose final judgment is
owned by an identifiable human reviewer who has checked the mathematics to
a depth adequate for that judgment. This designation does not give every
report the same evidentiary weight. Three authorship modes should be kept
distinct:

\begin{description}[
  leftmargin=2.8em,
  labelwidth=2.1em,
  style=nextline,
  itemsep=3pt,
  parsep=0pt
]

\item[H]
Human-authored review with no disclosed material
generative-AI contribution.

\item[HA]
Human-authored review with disclosed material AI assistance.

\item[AH]
AI-generated or AI-led analysis independently checked,
adopted, and owned by a human reviewer.

\end{description}

When the archive does not support an assignment, we leave the mode
unclassified. In particular, the absence of a disclosure is not sufficient
evidence from which to reconstruct an undocumented workflow.

\subsection{What an assessment can establish}

Four outcomes guide the reading of the reports. A \emph{confirmed
mathematical error} requires an established false statement, invalid
inference, violated hypothesis, or reproducible counterexample. An
\emph{unresolved proof gap} concerns a necessary step that has not been
justified sufficiently for independent verification. An \emph{expository
omission} is a compressed or missing derivation that can be recovered
without changing the claim. Finally, \emph{verification incomplete} means
that the checking was not deep enough to support either a positive or a
negative mathematical judgment about the relevant step or dependency.

These categories prevent a favorable result from saying more than the
review warrants. Failure to find an error does not establish that the whole
proof has been verified. The most useful reports therefore identify the
work left undone as carefully as the work completed. A Chapter~1 assessment,
for example, checks a central asymptotic constant and accepts the internal
consistency of the argument while declining to claim a fresh derivation of
every Mellin- and Fourier-analytic estimate.

There are also three objects to distinguish:
\[
T_{\mathrm{formal}},\qquad
T_{\mathrm{published}},\qquad
C_{\mathrm{advertised}}.
\]
They denote the proposition encoded in Lean, the theorem stated and argued
for in the manuscript, and the broader achievement attributed to it. The
first comparison concerns mathematical meaning:
\[
T_{\mathrm{formal}}
\longleftrightarrow
T_{\mathrm{published}}
\qquad\text{(semantic correspondence)}.
\]
The published theorem's relation to the advertised achievement requires a
separate assessment of scope, prior literature, novelty, significance, and
attribution. It should not be represented as a logical implication.
Neither task is settled automatically by the existence of a certificate.
This approach agrees with the Leiden Declaration's emphasis on transparency,
human responsibility, disclosure, and independent verifiability
\citep{Leiden2026}.

\section{Two cases that set the limits of the conclusion}

Across the 18 chapter-specific documents, no confirmed substantive
mathematical error in a principal result survives in the current review
record. Chapters~6 and 8 show why this conclusion needs its exact wording.
One concerns a criticism that had to be withdrawn; the other concerns
analytic work that a reviewer could not regard as fully checked.

\subsection{Chapter 6: a criticism of an altered formula}

An early audit of the quantum parallel-repetition argument identified what
appeared to be a polarity error in a greedy conditioning lemma. The issue
changed when the typeset manuscript was inspected directly. Its relevant
condition was
\[
\frac{1}{n-|D|}
\sum_{i\notin D}
\mathbb{P}(\overline{W_i}\mid W_D)
>
\delta.
\]
The probability concerns failure, represented by the complement
$\overline{W_i}$. Automatic PDF extraction had removed the overbar and
turned the expression into the corresponding success probability.

The resulting objection was mathematically plausible for the altered
formula, and could even support a counterexample to it. That formula,
however, was not the one in the manuscript. The revised public audit
withdraws the polarity claim and records no confirmed mathematical defect
in the examined argument \citep{SienickiSienicki2026}. An account that
continues to list this as a confirmed Chapter~6 error is out of date.

The incident exposes a failure in the review process itself. Between a
typeset proof and a reviewer's reasoning lies a chain of document handling
that can change mathematical meaning. Reviewing the original source is
therefore part of checking an objection, particularly when the objection
turns on a sign, complement, index, or hypothesis. The correction belongs
in the permanent record because it explains both the earlier judgment and
the reason for replacing it.

\subsection{Chapter 8: analytic details still needing scrutiny}

The two Chapter~8 assessments agree that no mathematical error has been
confirmed, but differ substantially over the readiness of the exposition.
One recommends minor revision. The other recommends major revision because
the treatment of uniform Bergman/Laplace asymptotics, the limiting
plurisubharmonic ray, slope passage, and the hypotheses needed for positivity
arguments is too compressed to sustain complete independent verification.

The critical assessment does not offer a counterexample. It identifies
transitions whose justification must be developed further for the argument
to be independently checked. In the terminology of this article, the
appropriate status is \emph{verification incomplete, with possible
unresolved proof gaps in the presentation}. Calling the result either
disproved or fully verified would lose this distinction. So would averaging
the two revision recommendations into one intermediate score.

Subsequent research leaves this reservation intact. Liu's work on the
equality case of Ehrhart's volume conjecture presents itself as a counterpart
to the inequality recently proved by OpenAI \citep{Liu2026}. It explicitly
discloses that the main result was obtained using generative AI, including
GPT-5.6 Sol, Fable~5, and Danus. The work is independent of the OpenAI
project and represents substantial complementary theorem-level uptake.
It does not verify the analytic proof of OpenAI's sharp inequality, and it
should not be described as purely human-independent verification.

Equality classification and proof of the inequality are separate
mathematical tasks. Progress on one does not independently settle the
analytic transitions questioned in the other. The favorable external uptake
and the more demanding review can therefore stand together. The remaining
uncertainty concerns the depth of verification, and deserves to stay visible.

\section{What the remaining chapter assessments support}

The other eight chapters also need to be read through the particular work
reported, rather than through a common label of approval. In the discussion
below, expressions of confidence describe reviewer judgments; they are not
calibrated numerical probabilities.

\subsection{Bounds and asymptotics: Chapters 1 and 2}

Chapter~1 receives high-confidence assessments of its main Cohn--Elkies
asymptotic result, with no confirmed substantive error. The checking covers
normalizations, signs, limiting relations, and the Mellin/Fourier structure.
An independent calculation in one report recovers
\[
\alpha_*=
\frac12\log_2\!\left(\frac{2\pi}{e}\right)
\approx 0.6044005443.
\]
Agreement on this constant is a specific and useful piece of evidence.
Its reach remains bounded by the report's explicit reservation: the entire
analytic core was not reconstructed from first principles, and the dense
estimates were not uniformly rederived line by line.

The Chapter~2 reports likewise find no error in the improved binary-code,
spherical-code, or packing bounds. They identify the representation-theoretic
interfaces and compactness arguments as suitable targets for further
specialist checking. The favorable judgment rests on the numerical and
structural claims that survived inspection, alongside an explicit account
of where additional expertise would improve the verification.

\subsection{Constructions and their later use: Chapters 3 and 4}

No substantive error is reported in Chapter~3's non-sofic-group construction.
One review places particular emphasis on matching conventions in imported
results and respecting the hierarchy of auxiliary parameters. These are
material qualifications in an argument with several nested layers of
$o(N)$ bookkeeping.

Later work gives this construction an additional kind of support.
Fournier-Facio's \emph{A Torsion-Free Non-Sofic Group} starts from OpenAI's
announcement and produces a different source of examples through the same
technical criterion \citep{FournierFacio2026}. Kun and Thom explicitly
build on the breakthrough, examine its mechanism, and derive new non-sofic
generalized wreath products \citep{KunThom2026}. Such work requires active
mathematical engagement with the construction. It supplies independent
reuse of the proof mechanism, although neither paper is a sentence-by-sentence
verification of Chapter~3.

Chapter~4 has one chapter-specific assessment. It judges the principal
construction sound and identifies mainly notational and expository issues,
with no substantive error found. The conclusion is favorable, but its review
redundancy is lower than that of several other chapters.

Zhou's independent and concurrent work constructs non-isomorphic ICC
property-(T) groups with isomorphic group von Neumann algebras, giving a
counterexample to Connes's rigidity conjecture \citep{Zhou2026}. Zhou
discloses assistance from GPT-5.6 Sol. The research output is independent
of the OpenAI project; that does not make it purely human-independent
corroboration.

The theorem's scope also matters. Zhou supplies a pair of such groups,
whereas OpenAI's principal formalized result asserts an infinite pairwise
non-isomorphic family with isomorphic factors \citep{OpenAI2026Lean}.
The follow-up confirms the conjecture-level conclusion that Connes's
rigidity conjecture is false. It does not independently reproduce the
stronger infinite-family theorem. The falsity of the conjecture would
therefore no longer depend solely on OpenAI's argument even if a local
flaw in that argument were later identified.

\subsection{Correctness and credit: Chapter 5}

The three Chapter~5 assessments are favorable about the permanent
lower-bound proofs and their parameter choices. They identify no confirmed
substantive error and regard the mathematics as highly likely to be correct.
Their main reservations concern how originality is described and credit
assigned.

Baur--Strassen differentiation and coefficient-counting frameworks were
already established tools. The more defensible novelty claim concerns the
permanent-specific constructions that make these tools effective at the
stated scale. This observation does not identify a defect in the theorem.
It asks the exposition to distinguish clearly between a new result and
a new general method. A review can support the proof while still asking
for a more careful account of its place in the literature.

\subsection{Hardness and parameter conversion: Chapter 7}

The Chapter~7 reviewers judge the reconstruction, valuation arguments,
approximation-factor conversion, and exponent bookkeeping coherent. They
find no confirmed substantive error and consider the principal result
highly likely to be correct. Some threshold and degree calculations remain
dense or abbreviated, so the strength of the assessment should be expressed
alongside that limitation.

Song's subsequent deterministic inapproximability results strengthen the
OpenAI bounds \citep{Song2026}. The Euclidean closest-vector factor
improves from $n^{1/400}$ to
\[
n^{1/8-\epsilon},
\]
and the binary nearest-codeword factor improves from $n^{1/200}$ to
\[
n^{1/4-\epsilon}.
\]
A stronger theorem is not a line-by-line audit of the earlier proof.
Nevertheless, provided the problem definitions, complexity assumptions,
and parameter conventions match, a valid stronger hardness theorem entails
the weaker hardness conclusion. Among the follow-ups considered here,
this is the strongest direct external corroboration of an OpenAI principal
conclusion at the theorem level.

\subsection{Recursive arguments and a missing calculation:
Chapters 9 and 10}

The sole Chapter~9 assessment finds the recursive coloring construction,
palette count, and final asymptotic conversion consistent. It reports no
confirmed error and judges the main theorem very likely correct. As in
Chapter~4, the limited number of assessments constrains how much review
redundancy can be claimed.

Both Chapter~10 reports assess the two main theorems favorably and find
no confirmed error. One calculation deserves a more explicit presentation:
an optimization in Lemma~5.1 is described as ``direct'' although the steps
should be shown. The appropriate classification is an expository omission
unless a later calculation establishes a mathematical problem. This keeps
the request for a fuller derivation separate from an unsupported allegation
of incorrectness.

\section{Public uptake and the weight it can bear}

The follow-up papers illustrate why external evidence needs to be described
by its relation to the original theorem. Mechanism reuse, a separate
counterexample, a stronger theorem, and a complementary result each add
something different. Research independent of the OpenAI project can also
involve AI assistance, as the disclosures by Zhou and Liu make clear.
Independence from the original proof-production project should therefore
be distinguished from independence from AI systems.

Reference works provide another form of evidence. In the public record
consulted for this article, Wolfram MathWorld incorporates results from
the OpenAI volume into entries on hypersphere packing, the MRRW bound,
the soficity conjecture, Connes's rigidity conjecture, the permanent,
quantum parallel repetition, the closest vector problem, Ehrhart's volume
conjecture, Ramsey numbers, Shannon capacity, and the Erd\H{o}s degeneracy
conjecture
\citep{
MathWorldSpherePacking2026,
MathWorldMRRW2026,
MathWorldSoficity2026,
MathWorldConnes2026,
MathWorldPermanent2026,
MathWorldQPR2026,
MathWorldCVP2026,
MathWorldEhrhart2026,
MathWorldRamsey2026,
MathWorldShannon2026,
MathWorldDegeneracy2026}.

This is reference-level recognition: the results are entering the ordinary
resources through which mathematicians locate and contextualize knowledge.
It is weaker evidence than active mathematical use of a proof mechanism,
and should not be presented as peer review or independent certification.
MathWorld's inclusion of a result does not establish how extensively its
proof has been checked.

The timing of an entry matters as well. The Ehrhart page accessed for this
article still listed the equality classification as open, whereas Liu's
August 2026 preprint claims to resolve it
\citep{MathWorldEhrhart2026,Liu2026}. This illustrates the different update
cycles of preprints and reference works. It is not a criticism of MathWorld,
but a reason to attach access dates to claims about public uptake and to
avoid treating a changing reference page as a fixed certification record.

\begin{table}[htbp]
\centering
\small

\caption{Forms of evidence and the questions they address. These are
distinct evidentiary acts, not a single strict ranking of verification.}

\label{tab:evidence}

\begin{tabularx}{\textwidth}{@{}p{0.26\textwidth}Y@{}}
\toprule
Evidence & Contribution and limit \\
\midrule

Formal certificate &
Checks the encoded proposition within the formal
system; correspondence to the published claim still needs assessment.
\\

Human technical review &
Examines the mathematics to a stated depth;
its conclusion must remain within the scope actually checked.
\\

Independent derivation &
Supplies another argument for a conclusion;
the conclusions must be compared at the same level of generality.
\\

Strengthening or extension &
Develops a result further; a stronger theorem
entails a weaker one only with the required matching assumptions.
\\

Complementary uptake &
Develops related mathematical content, such as an
equality case; it need not verify the original proof.
\\

Reference-work entry &
Documents recognition and contextualization;
it does not amount to independent certification.
\\

\bottomrule
\end{tabularx}
\end{table}

The case for confidence becomes stronger when these sources are read
together with their limits intact. It becomes less informative when they
are all reported as instances of a single, undefined act of verification.
The distinction is practical: a researcher deciding where to spend further
checking effort needs to know whether a theorem has been strengthened,
whether its mechanism has been reused, or whether its statement has simply
entered a reference resource.

\section{Making reviews open to correction}

Chapter~6 changes the familiar picture in which a human reviewer supplies
the final reliable check on an AI-produced proof. Human criticism can fail
too, including when the mathematics of the criticism is sound but its
source formula has been corrupted. A reliable process needs traceability
through the AI-generated argument, the formal object, the typeset
manuscript, extraction or processing, human review, and review of that
review.

Every transition has a possible failure mode. The generated reasoning can
be invalid; a formalization can encode the wrong proposition; extraction
can alter notation; a reviewer can misread an assumption; a reference
database can lag behind a new preprint. The response is to make the
connections inspectable. Public access to the human reasoning matters
alongside access to the theorem, since readers may need to understand why
an assessment changed.

The underlying reports and associated materials are deposited under
DOI \doi{10.5281/zenodo.22218229}
\citep{PeerReviewArchive2026}.\footnote{%
Authors of the discussed works and other interested researchers are welcome
to contribute comments and responses through the public
\emph{AI-Generated Science} Zenodo community:
\AIGSCommunityURL.}
This makes it possible to compare the synthesis with the primary review
record. A withdrawn objection should remain documented with the reason for
withdrawal, rather than disappear through an unexplained replacement.
Public deposition makes correction auditable; it does not make the reports
infallible.

The declared AIGS framework provides a useful basis for this practice.
Its scope, classification, and ethical requirements turn general approval
or disapproval into a more specific account of what the reviewer examined,
what assistance was used, and what uncertainty remains
\citep{AIGSScope2026,AIGSEthics2026}. The date qualification for the
15 August assessment form remains applicable: earlier reports should be
described by the requirements actually communicated to their authors.

The same reasoning applies to tools used in review. Mathematical work
already draws on symbolic algebra, search, proof assistants, databases,
and increasingly language models. The relevant questions are whether
material assistance is disclosed, whether the reviewer independently checks
the mathematics sufficiently for the stated conclusion, and whether that
reviewer accepts responsibility. These requirements are consistent with
the Leiden Declaration \citep{Leiden2026}; disclosure should make the
process more legible without pretending that all disclosed workflows carry
the same evidentiary weight.

Future archival versions would benefit from a standardized report header
and an immutable manifest. Together they should record stable identifiers,
the exact manuscript version and checksum, report date, reviewer status
and expertise, scope of checking, authorship mode where disclosed,
material AI or formal-tool use, the Lean commit where relevant, and
unresolved dependencies. Such documentation would allow the collection to
function more like a reproducible verification dataset. The chapter
inventory in \Cref{tab:coverage} is a document count; it does not replace
that more detailed record.

These practices also constrain the language of the synthesis. The corpus
does not support claims that humans have independently reproduced all ten
proofs, that Lean has verified every claim in the book, or that 18 documents
represent 18 independent reviewers. What it does support is a favorable
assessment with an explicit account of the parts still requiring work.
Chapter~8 remains the clearest priority for further analytic scrutiny,
while Chapter~6 demonstrates why an apparent defect needs equally careful
scrutiny before it is accepted.

\section{Implications for a broader scientific audit}

The immediate finding is substantial but bounded. Human and
human-accountable assessments cover every chapter and identify no surviving
confirmed substantive error in a principal result. Their checking differs
in depth, redundancy, authorship mode, and degree of independence. Several
difficult dependencies remain only partly reconstructed. Positive reviews
and unresolved verification work can coexist, just as provisional
mathematical acceptance can coexist with continuing efforts to clarify
details.

The public Lean repository and subsequent research make the overall
evidence stronger without removing those distinctions. Chapter~3's
mechanism has been independently used; Chapter~4's conjecture-level
conclusion has independent support; Chapter~7 has a stronger theorem;
and Chapter~8 has complementary equality-case work. Some follow-ups
disclose AI assistance. MathWorld supplies evidence of reference-level
uptake. Reading these contributions accurately preserves both the progress
and the remaining work.

This audit belongs to a wider program by Miko{\l}aj Sienicki and
Krzysztof Sienicki. In parallel with the mathematical volume, the authors
have prepared \emph{Ten Advances in Theoretical and Mathematical Physics},
\emph{Ten Advances in Theoretical and Mathematical Chemistry}, and
\emph{Ten Advances in Theoretical and Mathematical Biology}. Those projects
ask how difficult AI-generated or AI-assisted arguments can be made
transparent, checkable, and accountable in settings where formal deduction
usually cannot settle scientific validity on its own.

The additional questions depend on the discipline. Physics requires
attention to modelling assumptions and approximation regimes. Chemistry
brings electronic structure, many-body modelling, and the relation between
mathematical certification and computation. Biology places still greater
weight on empirical adequacy, mechanism, and scale. The purpose of these
projects is to examine how the audit framework must change when
mathematical validity is one part of scientific validity, rather than to
increase a count of claimed AI advances.

A related manifesto, \emph{Generative--Critical Method of Science (GCM):
Toward a New Methodology for AI-Assisted Discovery and Human Verification},
is at an advanced stage of development and is intended for publication
as soon as possible. It draws together the lessons of these projects:
AI can help generate conjectures, models, proofs, and candidate
explanations, while human researchers remain responsible for testing
assumptions, matching the work to the scientific problem, weighing evidence,
identifying failure modes, and accepting responsibility for the claims.

The GCM is proposed as an extension of the traditional scientific method
for research in which generation and criticism are distributed among
people, AI systems, formal tools, computational methods, and independent
review. The present case explains why criticism must include the assessment
process itself. A verification system earns confidence partly through its
ability to show what went wrong and how the record was repaired.

\section{Conclusion}

The assessments examined here give substantial support to the principal
results in OpenAI's \emph{Ten Advances in Mathematics and Theoretical
Computer Science}. No surviving confirmed substantive mathematical error
in those results is identified in this corpus. That statement leaves room
for unresolved proof gaps and incomplete verification, particularly at the
analytic transitions singled out in Chapter~8.

The public formalizations supply kernel-checkable proofs of the encoded
propositions; independent follow-ups supply mechanism reuse, conjecture-level
confirmation, stronger hardness results, and complementary mathematical
work. Reference entries document uptake. These contributions reinforce
the assessment when their logical roles and disclosed AI assistance remain
clear. None licenses a blanket claim that every informal proof has been
independently reconstructed by human reviewers.

The working model of confidence is cumulative:
\[
\boxed{
\begin{gathered}
\text{generation}
+\text{formal checking}
+\text{human reconstruction}
\\
{}+\text{independent reuse}
+\text{public correction}.
\end{gathered}
}
\]
The origin of a proof, whether human or artificial, cannot replace
examination of its content. Nor can acceptance of a related formal
statement settle every published claim. The Chapter~6 correction makes the
final term in this model concrete: human reviewers are essential, and
their judgments must remain open to inspection and revision alongside the
mathematics they assess.

\Needspace{12\baselineskip}
\section*{Materials and reproducibility}

The chapter reports, associated comments, and audit materials are publicly
archived on Zenodo under DOI \doi{10.5281/zenodo.22218229}. The DOI is the
stable archival identifier for the collection; the \emph{AI-Generated
Science} community provides its current browsing and discussion interface,
including comments, responses, and related analyses:

\begin{center}
\AIGSCommunityURL
\end{center}

The counts of 23 documents and 18 chapter-specific assessments describe
the snapshot examined on 27 August 2026. A dynamic community page is not
an immutable manifest. The archival release accompanying the article
should therefore include the per-document identifiers, dates, versions,
reviewer status, disclosed AI use, scope, and checksums described above.

\section*{Acknowledgments}

We thank F\'elix de la Poterie-Sienicki of McGill University,
Montr\'eal, Canada, for bringing the relevance of the OpenAI article to our
attention and for supporting this review project. We are grateful to the
researchers, most of them mathematicians, who gave their time and expertise
to examine difficult arguments, request clarification, challenge preliminary
conclusions, and contribute independent assessments. Comments shared both
publicly and privately, including those from contributors who preferred
to remain unnamed, improved the mathematical discussion and the review
methodology. Responsibility for remaining errors, omissions, and
interpretations rests with the authors.

\section*{AI assistance and responsibility}

Generative AI assisted the organization, writing, checking, and revision
of this article. The named authors retain responsibility for the
mathematical assessment, the accuracy of the references, and the final text.

\section*{Author involvement and conflicts of interest}

The authors organized and participated in the broader project and
contributed to the review record, including the Chapter~6 audit. They are
therefore not independent of the corpus synthesized here. This article
offers a methodological and evaluative synthesis, rather than external
institutional certification. Assessments supplied by independent
contributors are distinguished from the authors' synthesis, and the
underlying materials are public so that readers can examine that distinction
and the evidence for themselves. The article is not an official assessment
by OpenAI, Wolfram Research, arXiv, or the authors of the cited follow-up
papers.

\clearpage
\begingroup
\small


\endgroup
\end{document}